%% file: main.tex
\documentclass[10pt,twocolumn,letterpaper]{article}

\usepackage[pagenumbers]{wacv} 

\input{preamble}

\definecolor{wacvblue}{rgb}{0.21,0.49,0.74}
\usepackage[pagebackref,breaklinks,colorlinks,allcolors=wacvblue]{hyperref}
\usepackage{graphicx}
\usepackage{tikz}
\usetikzlibrary{arrows.meta, positioning, shapes.geometric, fit, backgrounds}
\definecolor{GeoBlue}{HTML}{0B4EA2}
\definecolor{GeoTeal}{HTML}{157A8A}
\definecolor{GeoGrey}{HTML}{F4F7FA}
\definecolor{GeoDark}{HTML}{243447}
\definecolor{GeoOrange}{HTML}{F28C28}
\definecolor{GeoRed}{HTML}{D9534F}
\definecolor{GeoGreen}{HTML}{2EAD6B}

\newcommand{\method}{GeoCond}
\usepackage{cuted}
\usepackage{capt-of}
\usepackage{comment}
\def\wacvPaperID{831} 
\def\confName{WACV}
\def\confYear{2027}

\title{\method: A Conditioning-Aware Reliability Adapter for \\ Feed-Forward 3D Reconstruction}

\author{David Ahmedt-Aristizabal$^{1}$, Mohammad Ali Armin$^{1}$, Russell Tsuchida$^{2}$, Lars Petersson$^{1}$ \\
$^{1}$ CSIRO, Australia, 
$^{2}$ Monash University, Australia\\ 
{\tt\small \{David.Ahmedtaristizabal,~Lars.Petersson\}@csiro.au,}
{\tt\small russell.tsuchida@monash.edu}
}

\begin{document}

\twocolumn[{%
\renewcommand\twocolumn[1][]{#1}%
\maketitle
\pagestyle{empty}
\begin{minipage}{\textwidth}
    \centering
    \vspace{-0.35cm}
    \includegraphics[width=0.65\linewidth]{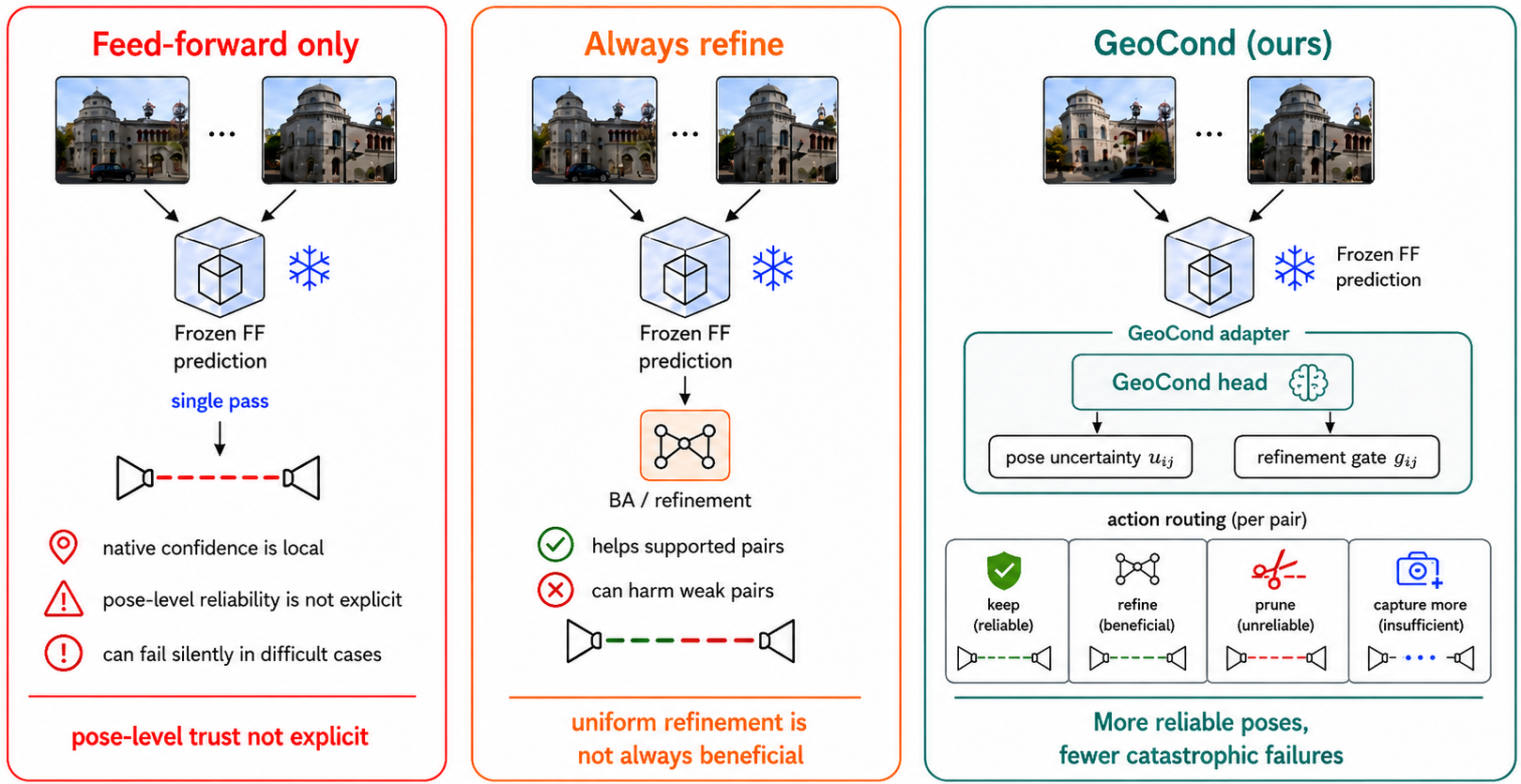}
    \vspace{-0.1cm}
    \captionof{figure}{
    \textbf{\method{} as a pose-level reliability adapter.} Feed-forward-only prediction provides no explicit pose-level trust signal and can fail silently in difficult cases, while uniformly applying bundle adjustment may help supported pairs but harm weak ones. \method{} augments a frozen feed-forward backbone with conditioning features to predict pose uncertainty and a refinement gate, selectively keeping, refining, pruning, or requesting additional views for each relative pose.
    }
    \label{fig:intro}
\end{minipage}

\vspace{3mm}    
}]

\input{sec/0_abstract}

\input{sec/1_intro}

\input{sec/2_related}

\input{sec/3_method}
\input{sec/4_experiments}
\input{sec/5_discussion_limitations}
\input{sec/6_conclusion}
\clearpage
{
    \small
    \bibliographystyle{ieeenat_fullname}
    \bibliography{main}
}
\appendix
\input{sec/supp}

\end{document}

%% file: preamble.tex

\makeatletter
\renewenvironment{abstract}{%
  \centerline{\large\bf Abstract}%
  \vspace*{3pt}%
  \it
}{%
}
\makeatother

%% file: sec/0_abstract.tex
\begin{abstract}
Feed-forward 3D foundation models such as VGGT predict cameras, depth, and point maps in a single pass, but can fail silently under low overlap, low parallax, and extreme relative rotation. 
Stratified analyses over these factors show that these failures are governed by geometric conditioning and are poorly captured by native aleatoric confidence. 
We introduce \textbf{\method{}}, a lightweight reliability adapter for frozen feed-forward 3D backbones. \method{} reads geometric conditioning cues and the backbone's own stability signals and outputs pose-level uncertainty and a refinement gate. During training, it can be supervised by frame-permutation orbit variance, ground-truth pose error when labels are available, or cycle residuals from unlabelled independent pose graphs. At inference, the default head requires only one backbone pass and a small MLP. 
On VGGT, \method{} improves out-of-distribution (OOD) AUSE (area under the sparsification-error curve; lower is better) from $0.32$ to $0.20$ over native confidence on held-out outdoor scenes (and to $0.21$ when trained on indoor data only) and avoids the collapse caused by applying bundle adjustment uniformly, without improving on feed-forward poses there. For three of six backbones, cycle-distilled variants provide a ground-truth-free adaptation route, including cases where permutation variance vanishes on equivariant models. The same reliability signal supports gated refinement, pose-graph weighting, calibration, curation, and capture decisions. Reliable feed-forward 3D reconstruction requires not only predicting geometry, but also knowing when that geometry should be trusted.
\href{https://geocond.github.io/}{[\textbf{Project website}]}
\vspace{-0.15cm}
\end{abstract}

%% file: sec/1_intro.tex
\section{Introduction}
\label{sec:intro}

Feed-forward 3D reconstruction models---from DUSt3R~\cite{dust3r} and MASt3R~\cite{mast3r} to VGGT~\cite{vggt} and recent successors~\cite{pi3,mapanything,vggtomega}---have made multi-view geometry increasingly accessible. Given one or more unposed images, they predict camera parameters, depth, dense point maps, and tracks in a single forward pass, often without the iterative optimization used in classical SfM/MVS pipelines~\cite{colmap,ba}. However, their reliability remains difficult to assess: a predicted pose can be wrong, yet the model provides no reliable pose-level signal that the output should not be trusted.

The missing layer is pose-level reliability. Feed-forward 3D backbones emit per-pixel, per-point, or per-track confidence, but these signals do not answer the downstream question: should this relative camera pose be trusted? A pair can contain many confident local predictions and still be globally ill-conditioned if the views barely overlap, the camera motion has little triangulation parallax, or a large relative rotation leaves weak correspondences. In these regimes, the pose can fail while native confidence remains weakly informative.

Geometric conditioning also determines whether refinement is beneficial. Bundle adjustment (BA) is often used as a post-processing step for feed-forward estimates, and learned or test-time refinement has been explored in related pipelines~\cite{vggsfm,selfi}. On a frozen VGGT, BA can improve pairs with sufficient track support but corrupt low-overlap pairs whose geometric constraints are weak. Thus, the relevant question is not only how to refine a feed-forward prediction, but whether refinement should be applied at all.

We address this problem with \textbf{\method{}}, a lightweight conditioning-aware reliability adapter attached to a frozen feed-forward 3D backbone (Figure~\ref{fig:intro}). 
\method{} uses seven interpretable features computed from the backbone's own predictions: four geometric conditioning cues (predicted parallax, co-visibility, baseline, relative rotation) and three backbone stability signals (camera-head rotation and translation correction and aggregated aleatoric confidence). 
During training, its uncertainty output can be supervised by frame-permutation orbit variance, target-domain pose error when labels exist, or cycle residuals from unlabelled independent pose graphs; the gate needs labelled pairs.
At inference, the default head requires only one backbone pass and produces a pose-level uncertainty score and a refinement gate for each relative pose.
Our contributions are:
\begin{enumerate}
    \item We identify geometric conditioning as a primary driver of feed-forward relative-pose failure, with direct evidence for low parallax, low overlap, and large relative rotation on indoor and outdoor data, and show that native aleatoric confidence is a poor predictor of these pose-level errors.
    \item We introduce \textbf{\method{}}, a lightweight reliability adapter that predicts pose-level uncertainty and refinement decisions from a frozen backbone's own predicted geometry and analyse which inputs carry the reliability signal.
    \item We propose three reliability-supervision settings: frame-permutation distillation, supervised target-domain training, and ground-truth-free cycle-residual distillation.
    \item We show that refinement is conditioning-dependent: BA can help well-supported pairs but harm low-overlap cases, motivating a learned gate rather than uniform refinement.
    \item We demonstrate that the predicted reliability signal transfers zero-shot from indoor scenes to out-of-distribution (OOD) outdoor extreme-view data, can be adapted to other backbones with backbone-dependent gains, and supports downstream pose-graph weighting, calibration, curation, and adaptive capture.
\end{enumerate}

%% file: sec/2_related.tex
\section{Related work}
\label{sec:related_work}

\noindent\textbf{Feed-forward 3D reconstruction.}
Classical SfM/MVS estimates camera motion and structure through matching, triangulation, and bundle adjustment~\cite{colmap,ba}. Feed-forward models instead predict geometry directly. DUSt3R~\cite{dust3r} and MASt3R~\cite{mast3r} regress point maps, while later systems extend this paradigm to spatial memory, many-view inference, streaming, and long sequences~\cite{spann3r,fast3r,streamvggt,cut3r,vggtlong}. VGGT~\cite{vggt} predicts cameras, depth, point maps, and tracks, with extensions for permutation equivariance, metric scale, efficiency, and dense SLAM~\cite{pi3,mapanything,vggtomega,mast3rslam}. Despite these advances, existing models do not expose a pose-level reliability signal for deciding whether a relative pose should be trusted.

\noindent\textbf{Robustness of feed-forward 3D.}
Recent analyses study failure modes of feed-forward 3D models, including distractor-view rejection~\cite{robustvggt}, emergent epipolar geometry in intermediate layers~\cite{geomprior}, attention degradation with sequence length~\cite{attncollapse}, and catastrophic errors under minimal-overlap extreme views~\cite{extremeview}. Our work targets this regime directly, but rather than filtering views or modifying the backbone, we predict pose uncertainty and use it to abstain, down-weight, refine, or adapt unreliable relative poses.

\noindent\textbf{Uncertainty for 3D regression.} 
Deep uncertainty is commonly separated into aleatoric and epistemic components~\cite{kendall2017uncertainties}, with MC dropout, ensembles, and evidential learning as standard estimators~\cite{gal2016dropout,deepensembles,sensoy2018evidential}. Feed-forward 3D backbones typically output per-pixel or per-point confidence, while Trust3R~\cite{trust3r} models evidential uncertainty for dense geometry. Our target is different: pose-level reliability under geometric degeneracy, including low overlap, low parallax, and extreme relative rotation. We supervise it using frame-permutation orbit variance, pose error, or cycle residuals from unlabelled pose graphs.

\begin{figure*}[!t]
    \centering
    \includegraphics[width=0.75\linewidth]{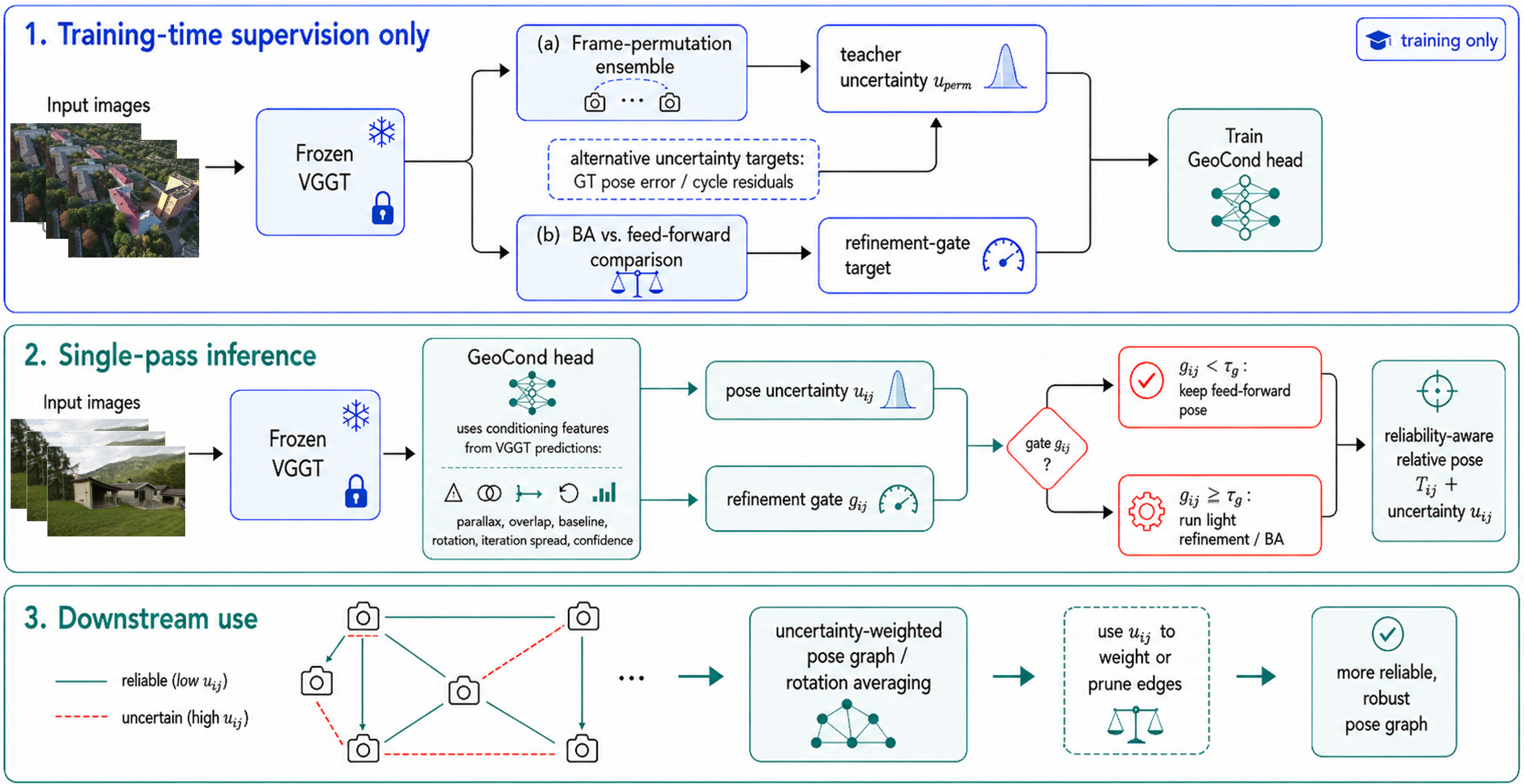}
    \caption{\textbf{Overview of \method{}.}
    In the default VGGT setting, training uses a frozen backbone to generate an uncertainty teacher from frame-permutation orbit variance and a refinement target from BA-vs-feed-forward comparison. More generally, the reliability teacher can also come from target-domain pose errors or cycle residuals from unlabeled pose graphs. At inference, the backbone runs once and a lightweight GeoCond head predicts pose uncertainty $u_{ij}$ and a refinement gate $g_{ij}$ from conditioning features derived from the backbone's own predictions. The predicted uncertainty can also weight or prune edges in a downstream pose graph.
    }
    \label{fig:overview}
\vspace{-0.3cm}
\end{figure*}

\noindent\textbf{Refinement and pose-graph consistency.}
Bundle adjustment remains central to classical reconstruction~\cite{ba}, and recent learned systems incorporate or approximate geometric refinement through differentiable BA, recurrent dense BA, relative-pose regression, geometric-consistency distillation, or test-time refinement~\cite{vggsfm,droidslam,reloc3r,geloc3r,selfi}. 
These methods improve or amortize estimation; we instead show that refinement itself is conditioning-dependent and should be gated rather than applied uniformly. 
Classical SfM also uses loop constraints, rotation averaging, and robust synchronization to reject inconsistent edges~\cite{rotavg,lerman2022robust,zach2010disambiguating}. We use the same geometric principle to turn cycle residuals into an unlabelled teacher for a deployable feed-forward reliability head.



%% file: sec/3_method.tex
\section{Method}
\label{sec:method}


\subsection{Overview}
\label{sec:method_overview}

Our goal is to augment a frozen feed-forward 3D backbone with a lightweight reliability layer. We introduce \method{}, a conditioning-aware head that predicts pose-level uncertainty and decides whether geometric refinement should be applied. Figure~\ref{fig:overview} summarizes the method.

Training and inference are separated. In the default VGGT setting, the frozen backbone provides two supervision signals: a frame-permutation ensemble, which yields an uncertainty teacher, and a comparison between bundle-adjusted and feed-forward poses, which provides the target for a refinement gate and requires ground-truth poses. More generally, the uncertainty target can also come from target-domain pose errors when labels are available, or from cycle residuals in unlabelled independent pose graphs. In all cases, the backbone remains frozen and only the \method{} head is optimised.

At inference, the backbone runs once.  The head outputs a pose-level uncertainty score $u_{ij}$ and a refinement gate $g_{ij}$ for each image pair. The gate determines whether to keep the feed-forward pose or apply light geometric refinement, while the uncertainty can be used to rank, prune, calibrate, or weight relative poses in downstream systems such as pose-graph fusion.

\subsection{Setup and pose error}
\label{sec:setup}
Let $\mathcal{I}=\{I_1,\ldots,I_S\}$ be a set of $S$ input images. A frozen feed-forward 3D backbone predicts per-frame extrinsics $E_i$, intrinsics $K_i$, depth maps $D_i$, point maps, native confidence maps $A_i$, and optionally camera tokens $z_i$. We denote by $\hat{T}_i\in SE(3)$ the predicted world-to-camera pose and by $\hat{\mathbf{c}}_i$ its camera centre.

For an image pair $(i,j)$, the feed-forward relative pose is
\vspace{-0.2cm}
\begin{equation}
\hat{T}^{\mathrm{FF}}_{ij}
=
\hat{T}_{j}\hat{T}^{-1}_{i}
=
(\hat{R}_{ij},\hat{\mathbf{t}}_{ij}),
\end{equation}
where $\hat{R}_{ij}\in SO(3)$ is the relative rotation and $\hat{\mathbf{t}}_{ij}$ denotes the relative translation direction.  Given the ground-truth relative pose $T^{\star}_{ij}=(R^{\star}_{ij},\mathbf{t}^{\star}_{ij})$, we define the pairwise pose error as
\vspace{-0.2cm}
\begin{equation}
e_{ij}
=
\max
\left(
d_R(\hat{R}_{ij},R^{\star}_{ij}),
d_t(\hat{\mathbf{t}}_{ij},\mathbf{t}^{\star}_{ij})
\right),
\end{equation}
where
\vspace{-0.2cm}
\begin{equation}
\begin{aligned}
d_R(R_1,R_2)
&=
\cos^{-1}\!\left(
\frac{\operatorname{tr}(R_1^\top R_2)-1}{2}
\right),\\
d_t(\mathbf{t}_1,\mathbf{t}_2)
&=
\cos^{-1}\!\left(
\frac{\mathbf{t}_1^\top \mathbf{t}_2}
{\lVert \mathbf{t}_1\rVert_2 \lVert \mathbf{t}_2\rVert_2}
\right).
\end{aligned}
\end{equation}
As in prior work~\cite{vggsfm}, we report relative-pose AUC@$\tau$ over thresholds $\tau\in\{5,10,20,30\}$ degrees.  AUSE is defined in Sec.~\ref{sec:experiments}.

\subsection{Conditioning features}
\label{sec:features}
Our central hypothesis is that feed-forward pose error is governed by geometric conditioning. We therefore construct a low-dimensional per-pair conditioning vector from a single frozen forward pass, combining geometric cues with backbone stability signals:
\begin{equation}
\boldsymbol{\phi}_{ij}
=
[
\hat{\alpha}_{ij},
\hat{\rho}_{ij},
\hat{b}_{ij},
\hat{\theta}_{ij},
\hat{s}^{R}_{ij},
\hat{s}^{t}_{ij},
\hat{a}_{ij}
],
\end{equation}
where $\hat{\alpha}_{ij}$ is predicted parallax, $\hat{\rho}_{ij}$ is predicted co-visibility or overlap, $\hat{b}_{ij}=\lVert \hat{\mathbf{c}}_i-\hat{\mathbf{c}}_j\rVert_2$ is predicted baseline, $\hat{\theta}_{ij}=d_R(\hat{R}_{ij},I)$ is relative rotation magnitude, $\hat{s}^{R}_{ij}$ and $\hat{s}^{t}_{ij}$ measure the first-to-last correction made by the backbone's iterative camera head, for rotation and translation respectively, and $\hat{a}_{ij}$ is the negated mean per-pixel depth confidence over the two frames, so larger values indicate lower native confidence.

We compute parallax as the median triangulation angle induced by the predicted depths, intrinsics, and camera poses. Co-visibility is estimated from the fraction of mutually valid projected evidence between the two predicted views, using the backbone's predicted geometry and validity/confidence masks. 
The first four scalars are geometric conditioning cues: they describe regimes where relative pose is poorly conditioned, such as low parallax, low overlap, small baseline, or extreme rotation, and explain where failures concentrate (Sec.~\ref{sec:exp_law}). The last three are backbone-specific stability signals: they indicate whether a particular prediction is unstable and, as the ablation in Sec.~\ref{sec:exp_uq} shows, carry most of the ranking, with the geometric cues adding a consistent margin. Unless otherwise stated, this scalars-only representation is the default input to \method{}. We evaluate optional camera-token variants as an ablation, but find that token-based variants are more prone to domain overfitting.

\subsection{Reliability supervision signals}
\label{sec:teachers}

\method{} can be trained from different reliability signals depending on the available supervision. In the default VGGT setting, we use frame-permutation orbit variance as a training-time epistemic teacher. When labelled target-domain poses are available, the same head can be supervised directly from pose error or failure labels. When no ground truth is available, cycle residuals from independent pairwise pose graphs provide a self-supervised reliability signal. 
These alternatives use the same \method{} architecture and differ only in the target used to supervise the uncertainty output; we call the resulting heads \method{}-O (orbit-distilled, the default), \method{}-E (error-supervised), and \method{}-C (cycle-distilled).

\noindent\textbf{Frame-permutation orbit variance.}
For non-equivariant multi-view backbones such as VGGT, image ordering is a nuisance variable: the relative pose should not change when the same image set is re-ordered. Let $\pi_k$, $k=1,\ldots,K$, denote $K$ random permutations of the input set. For each permutation, we run the frozen backbone and recover
\vspace{-0.2cm}
\begin{equation}
\hat{T}^{(k)}_{ij}
=
\hat{T}^{(k)}_{j}
\left(\hat{T}^{(k)}_{i}\right)^{-1}.
\end{equation}
We define the orbit-variance teacher as the average angular spread of translation directions:
\vspace{-0.2cm}
\begin{equation}
u^{\mathrm{perm}}_{ij}
=
\frac{2}{K(K-1)}
\sum_{a<b}
d_t
\left(
\hat{\mathbf{t}}^{(a)}_{ij},
\hat{\mathbf{t}}^{(b)}_{ij}
\right).
\vspace{-0.2cm}
\end{equation}
If the relative pose changes substantially under harmless re-orderings of the same input set, the backbone is unstable for that pair. This teacher requires $K$ backbone passes only during training; inference is single-pass. The teacher is intentionally backbone-specific: for permutation-equivariant models such as $\pi^3$~\cite{pi3}, orbit variance can vanish.

\noindent\textbf{Ground-truth supervision.}
When labelled target-domain poses exist, the same head can be trained directly from pose error $e_{ij}$ or from binary failure labels such as $\mathbf{1}[e_{ij}>\tau]$.

\noindent\textbf{Cycle-residual distillation.}
When no labels exist, independent pairwise estimates form pose graphs whose loops should close. For a triangle $(i,j,k)$, the composed rotation should satisfy
\begin{equation}
R_{ij}R_{jk}R_{ki}=I.
\end{equation}
We define a cycle residual for an edge by averaging $\angle(R_{ij}R_{jk}R_{ki},I)$ over triangles incident to that edge. This residual requires no ground truth and only small matrix multiplications after independent edges are estimated. We use it either as a direct reliability score or as a teacher for a distilled head. This provides a ground-truth-free adaptation setting that also applies when the permutation teacher is uninformative for equivariant backbones.

\subsection{GeoCond head}
\label{sec:head}

\method{} is a small MLP applied independently to each image pair. Its default input is the standardized conditioning vector $\boldsymbol{\phi}_{ij}$ together with a compact representation of the current feed-forward relative pose,
\vspace{-0.2cm}
\begin{equation}
\psi(\hat{T}^{\mathrm{FF}}_{ij})
=
(q_{ij},\hat{\mathbf{t}}_{ij}),
\end{equation}
where $q_{ij}$ is the relative rotation represented as a unit quaternion and $\hat{\mathbf{t}}_{ij}$ is the unit translation direction. We use the scalars-only conditioning features by default and evaluate camera-token variants as an ablation.
Let
\begin{equation}
\mathbf{h}_{ij}
=
f_{\theta}
\left(
\boldsymbol{\phi}_{ij},
\psi(\hat{T}^{\mathrm{FF}}_{ij})
\right)
\end{equation}
be the shared pair representation produced by the GeoCond MLP. The head predicts three quantities:
\begin{equation}
\hat{u}_{ij}=h_u(\mathbf{h}_{ij}), \enspace
\hat{g}_{ij}=\sigma(h_g(\mathbf{h}_{ij})), \enspace
\hat{r}_{ij}=h_r(\mathbf{h}_{ij}),
\end{equation}
where $\hat{u}_{ij}$ is a pose-level uncertainty score, $\hat{g}_{ij}\in[0,1]$ is a refinement gate, and $\hat{r}_{ij}$ parameterizes an optional feed-forward SE(3) residual. The residual branch is optional and is used only to test whether geometric refinement can be amortized into a single forward pass.

The uncertainty output is trained from a reliability target $y^u_{ij}$. For continuous targets, such as log orbit variance, pose error, or cycle residuals, we use
\vspace{-0.2cm}
\begin{equation}
\mathcal{L}_{u}
=
\left(
\hat{u}_{ij}
-
y^u_{ij}
\right)^2.
\end{equation}
In the default VGGT setting,
\vspace{-0.2cm}
\begin{equation}
y^u_{ij}
=
\log\left(1+u^{\mathrm{perm}}_{ij}\right).
\end{equation}
When training from ground-truth pose error or cycle residuals, $y^u_{ij}$ is replaced by the corresponding normalized reliability target. For binary failure targets, such as $\mathbf{1}[e_{ij}>\tau]$, we replace the squared loss with a binary cross-entropy loss.

The refinement gate predicts whether BA improves the feed-forward pose. Let $e^{\mathrm{FF}}_{ij}$ and $e^{\mathrm{BA}}_{ij}$ be the pose errors before and after BA during training. The gate target is
\begin{equation}
y^g_{ij}
=
\mathbf{1}
\left[
e^{\mathrm{BA}}_{ij}
<
e^{\mathrm{FF}}_{ij}
\right],
\end{equation}
with loss
\begin{equation}
\mathcal{L}_{g}
=
\operatorname{BCE}
\left(
\hat{g}_{ij},
y^g_{ij}
\right).
\end{equation}
Unlike the uncertainty targets, $y^g_{ij}$ requires ground-truth poses, so the gate is trained on labelled pairs and transferred; Sec.~\ref{sec:exp_vggt} shows that it transfers only to domains where BA behaves similarly.

The optional residual branch is supervised by the better of the feed-forward and BA estimates:
\begin{equation}
T^{\star}_{ij}
=
\begin{cases}
T^{\mathrm{BA}}_{ij}, & e^{\mathrm{BA}}_{ij}<e^{\mathrm{FF}}_{ij},\\
T^{\mathrm{FF}}_{ij}, & \text{otherwise}.
\end{cases}
\end{equation}
Writing this target as $(q^{\star}_{ij},\hat{\mathbf{t}}^{\star}_{ij})$, the residual branch predicts a corrected pose $(\tilde{q}_{ij},\tilde{\mathbf{t}}_{ij})$ and is trained with
\begin{equation}
\mathcal{L}_{r}
=
\left(
1-\left|\tilde{q}_{ij}^{\top}q^{\star}_{ij}\right|
\right)
+
\left(
1-
\tilde{\mathbf{t}}_{ij}^{\top}
\hat{\mathbf{t}}^{\star}_{ij}
\right).
\vspace{-0.2cm}
\end{equation}
Quaternions and translation directions are normalized before computing the residual loss.

The full training objective is
\vspace{-0.2cm}
\begin{equation}
\mathcal{L}
=
\mathcal{L}_{u}
+
\mathcal{L}_{g}
+
\lambda_r\mathcal{L}_{r},
\vspace{-0.2cm}
\end{equation}
with $\lambda_r=2$ when the residual branch is enabled and $\lambda_r=0$ otherwise. All backbone parameters remain frozen; only the \method{} head is optimized.

\subsection{Deployment decisions}
\label{sec:deployment_method}

At inference, \method{} adds only a small MLP on top of the frozen backbone: 19.6k parameters and $0.17$\,ms per ten pairs ($0.2\%$ of a $95.6$\,ms backbone pass). The seven conditioning scalars take $3.5$\,ms per five-frame tuple in a batched GPU implementation; the $K{=}6$ teacher is training-only, while one BA call costs $4.0$\,s. We use its outputs in four ways. First, for refinement control, the final relative pose is
\vspace{-0.2cm}
\begin{equation}
T^{\mathrm{out}}_{ij}
=
\begin{cases}
T^{\mathrm{ref}}_{ij}, & \hat{g}_{ij}\geq \tau_g,\\
T^{\mathrm{FF}}_{ij}, & \text{otherwise},
\end{cases}
\end{equation}
where $T^{\mathrm{ref}}_{ij}$ is obtained either from BA or from the optional residual branch. Second, for selective prediction, high-uncertainty pairs can be abstained from or pruned. Third, for pose-graph fusion, we weight each edge by
\vspace{-0.2cm}
\begin{equation}
w_{ij}=\exp(-\gamma \hat{u}_{ij})
\end{equation}
before rotation averaging:
\vspace{-0.2cm}
\begin{equation}
\min_{\{R_i\}_{i\in\mathcal{V}}}
\sum_{(i,j)\in\mathcal{E}}
w_{ij}
\,d_R
\left(
R_j R_i^{-1},
\hat{R}_{ij}
\right)^2.
\vspace{-0.2cm}
\end{equation}
Finally, for risk-controlled decisions, we apply Platt scaling~\cite{platt1999probabilistic} to map $\hat{u}_{ij}$ into a catastrophic-failure probability $\mathbb{P}(e_{ij}>\tau)$ and use calibrated thresholds for pseudo-label curation and adaptive capture.

%% file: sec/4_experiments.tex
\section{Experiments}
\label{sec:experiments}



\noindent\textbf{Datasets and protocol.}
The core experiments use frozen VGGT-1B. Our main indoor benchmark is 7-Scenes~\cite{sevenscenes}; our out-of-distribution (OOD) benchmark is MegaUnScene~\cite{extremeview}, an outdoor extreme-view benchmark with low overlap and large rotations. Additional tests use MegaDepth-1500, ScanNet-1500, and multiple feed-forward backbones: $\pi^3$~\cite{pi3}, DUSt3R~\cite{dust3r}, Fast3R~\cite{fast3r}, CUT3R~\cite{cut3r}, and MapAnything~\cite{mapanything}. Unless stated otherwise, heads are evaluated scene-disjoint from their training data. 
We report AUSE for uncertainty ranking and AUC@30 for pose accuracy or retained-pose quality. AUSE is the normalized area between the sparsification curve $S_u(f)$, obtained by removing the fraction $f$ of pairs with highest predicted uncertainty, and the oracle curve $S_o(f)$, obtained by removing pairs by true error:
$\int_0^{0.9}[S_u(f)-S_o(f)]\,\mathrm{d}f/S_u(0)$; lower is better. Training sets are 7S (7-Scenes training scenes only, i.e.\ strict zero-shot to MegaUnScene) or mix (7-Scenes plus MegaUnScene training scenes); test scenes are always held out. AUC@$ \tau $ is the mean fraction of pairs with $e_{ij}\le t$ over integer thresholds $t\le\tau$ degrees, times 100~\cite{vggsfm}.

\begin{figure*}[!t]
\centering
\includegraphics[width=0.36\linewidth]{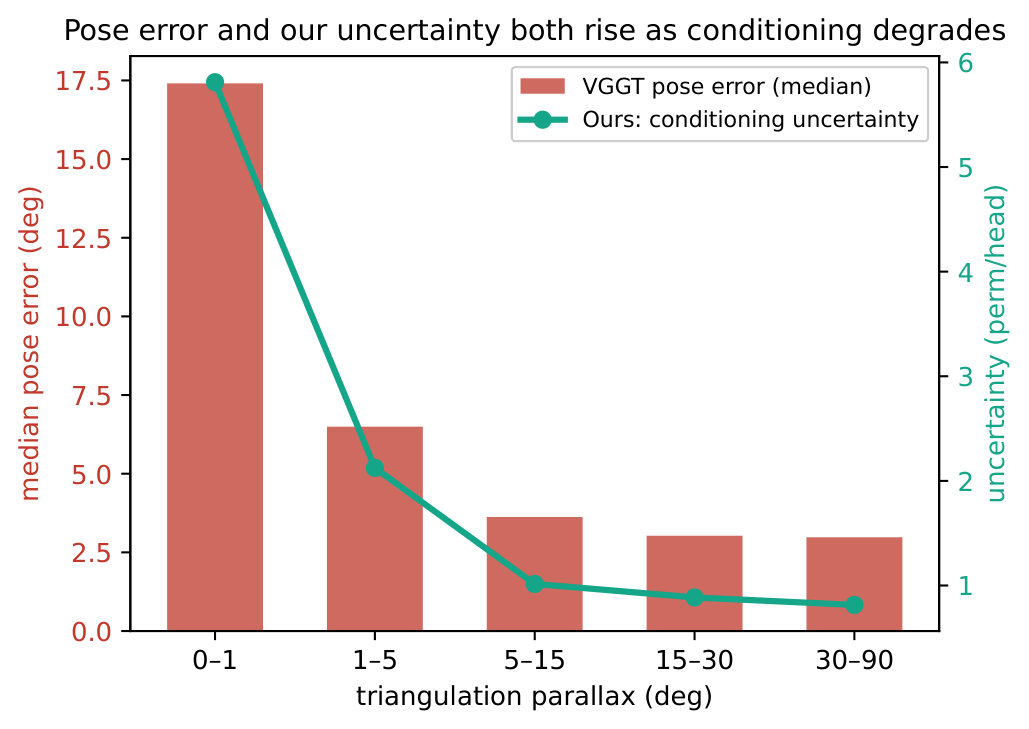}
\includegraphics[width=0.62\linewidth]{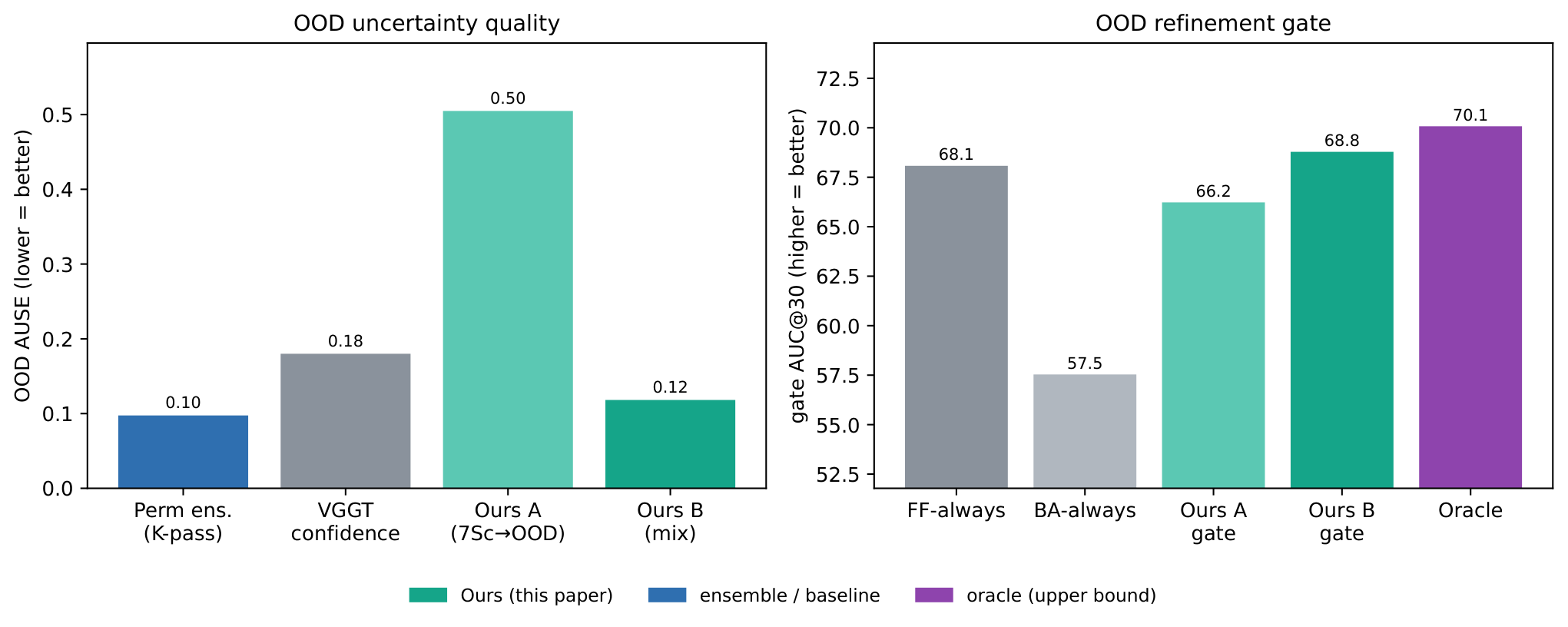}
\vspace{-0.3cm}
\caption{\textbf{Conditioning explains failures and decisions.}
\textit{Left:} pose error and orbit-variance uncertainty rise as parallax degrades, while native confidence remains weakly informative. 
\textit{Right:} on held-out OOD scenes (pairs of Table~\ref{tab:headline}, five seeds), \method{}-O trained on 7-Scenes only (7S) or on the mix ranks pose error far better than native confidence, but only the mix-trained gate avoids most of the damage of applying BA uniformly.
}
\label{fig:law}
\vspace{-0.3cm}
\end{figure*}

\subsection{Geometric conditioning explains pose failure}
\label{sec:exp_law}
On 7-Scenes, pose error is strongly parallax-dependent: the median pairwise error increases from approximately $3.0^\circ$ at high parallax to $17.4^\circ$ below $1^\circ$ parallax, with $22\%$ of low-parallax pairs becoming catastrophic failures above $30^\circ$. The frame-permutation orbit variance tracks this degradation, while VGGT's native aleatoric confidence remains weakly informative.


Table~\ref{tab:law} shows the same trend across indoor, outdoor extreme-view, and standard two-view benchmarks. Where parallax buckets are available, low-parallax pairs have substantially higher median error than high-parallax pairs. On MegaUnScene, for example, the median error rises from $3.3^\circ$ to $44.2^\circ$, while conditioning-based AUSE is $0.163$ compared with $0.323$ for native confidence. Figure~\ref{fig:law} visualizes this behaviour. These results support the central claim that pose reliability is governed by geometric conditioning rather than per-pixel confidence alone.
Table~\ref{tab:strat} further stratifies the analysis by overlap and relative rotation. Low-overlap and high-rotation pairs are substantially more failure-prone: in 7-Scenes office, overlap $<0.01$ gives median error $42.2^\circ$ with $54\%$ catastrophic pairs, and relative rotation $\geq90^\circ$ gives $85.5^\circ$ with $63\%$ catastrophic pairs. On MegaUnScene, the corresponding catastrophic rates are $32\%$ and $40\%$. OOD, \method{} improves over native confidence in every bin, including low overlap ($0.251$ vs.\ $0.429$ AUSE) and high rotation ($0.357$ vs.\ $0.547$ AUSE).

\begin{table}[!t]
\centering
\small
\caption{Geometric conditioning predicts pose error across datasets. Low/high parallax columns show median pose error in degrees; AUSE $\downarrow$ compares conditioning-based uncertainty with native VGGT confidence; the better of the two is in bold}.
\label{tab:law}
\resizebox{0.98\linewidth}{!}{
\begin{tabular}{lccccc}
\toprule
Dataset & $n$ & err low par. & err high par. & cond. AUSE & native AUSE \\
\midrule
7-Scenes heads & 2200 & 17.4 & 3.0 & \textbf{0.172} & 0.417 \\
7-Scenes chess & 1600 & 10.8 & 2.7 & \textbf{0.157} & 0.370 \\
7-Scenes office & 700 & 22.6 & 10.0 & \textbf{0.173} & 0.186 \\
MegaUnScene & 900 & 44.2 & 3.3 & \textbf{0.163} & 0.323 \\
MegaDepth-1500 & 1500 & -- & -- & \textbf{0.261} & 0.616 \\
ScanNet-1500 & 1500 & 8.6 & 2.2 & \textbf{0.230} & 0.386 \\
\bottomrule
\end{tabular}}
\vspace{-0.3cm}
\end{table}

\subsection{Single-pass uncertainty and gated refinement}
\label{sec:exp_vggt}


We evaluate three VGGT settings: 
\textit{in-domain distillation}, where a single-pass head is trained to match the orbit-variance teacher on held-out 7-Scenes tuples; \textit{strict zero-shot OOD transfer} (7S), where the head is trained only on 7-Scenes and tested directly on MegaUnScene; and 
a \textit{mixed-domain head (mix)}, trained on a 7-Scenes+MegaUnScene mix and evaluated on held-out indoor and outdoor scenes for the main baseline comparison.

\noindent\textbf{In-domain distillation.}
On held-out 7-Scenes tuples, \method{} distils the $K$-pass orbit teacher into one forward pass: AUSE $0.24$ versus $0.22$ for the teacher and $0.44$ for native confidence. Its gated refinement reaches $85.6$ AUC@30, outperforming feed-forward-only ($82.1$) and BA-always ($84.1$), with an oracle of $88.4$. A scalars-only head performs at least as well (AUSE $0.20$, gated AUC@30 $86.1$), making it our default model.

\noindent\textbf{Strict zero-shot OOD transfer (7S).}
On MegaUnScene, the same conditioning principle holds under extreme viewpoint changes (maximum rotation $173.7^\circ$, $41\%$ zero-overlap pairs, and $22\%$ catastrophic failures on the held-out scenes). Trained on 7-Scenes only, \method{}-O (7S) reaches AUSE $0.216{\pm}0.015$ on these scenes (five seeds), close to mixed training ($0.209{\pm}0.004$) and far better than native confidence ($0.323$), whereas the camera-token variant does not transfer ($0.361$). The gate does not transfer: trained where BA helps $40\%$ of pairs, it requests BA for $70\%$ of OOD pairs and reaches $46.0$ AUC@30, between BA-always ($38.4$) and feed-forward-only ($60.3$); mixed training raises it to $57.6$ (Figure~\ref{fig:law}, right).

\begin{table}[!t]
\centering
\small
\caption{Pose error and uncertainty quality stratified by ground-truth overlap and relative rotation on held-out 7-Scenes office and OOD MegaUnScene pairs. AUSE $\downarrow$ compares native confidence and \method{} within each bin; the better value is bold.}
\label{tab:strat}
\vspace{-0.1cm}
\resizebox{0.98\linewidth}{!}{
\begin{tabular}{llcccc}
\toprule
Dataset & bin & $n$ & med. err & cat. rate & native / \method{} AUSE \\
\midrule
7-Scenes office & overlap $<0.01$ & 227 & 42.2 & 54\% & \textbf{0.263} / 0.284 \\
7-Scenes office & overlap $0.01$--$0.3$ & 112 & 6.8 & 11\% & 0.189 / \textbf{0.064} \\
7-Scenes office & overlap $\geq0.3$ & 361 & 7.4 & 19\% & 0.213 / \textbf{0.142} \\
7-Scenes office & rel. rotation $<30^\circ$ & 362 & 7.6 & 19\% & 0.207 / \textbf{0.139} \\
7-Scenes office & rel. rotation $30^\circ$--$90^\circ$ & 179 & 11.0 & 20\% & 0.231 / \textbf{0.141} \\
7-Scenes office & rel. rotation $\geq90^\circ$ & 159 & 85.5 & 63\% & \textbf{0.178} / 0.204 \\
\midrule
MegaUnScene (OOD) & overlap $<0.01$ & 491 & 11.3 & 32\% & 0.429 / \textbf{0.251} \\
MegaUnScene (OOD) & overlap $0.01$--$0.3$ & 305 & 3.5 & 5\% & 0.193 / \textbf{0.165} \\
MegaUnScene (OOD) & overlap $\geq0.3$ & 104 & 10.0 & 27\% & 0.385 / \textbf{0.285} \\
MegaUnScene (OOD) & rel. rotation $<30^\circ$ & 323 & 5.7 & 14\% & 0.333 / \textbf{0.207} \\
MegaUnScene (OOD) & rel. rotation $30^\circ$--$90^\circ$ & 400 & 5.2 & 21\% & 0.239 / \textbf{0.149} \\
MegaUnScene (OOD) & rel. rotation $\geq90^\circ$ & 177 & 18.7 & 40\% & 0.547 / \textbf{0.357} \\
\bottomrule
\end{tabular}}
\vspace{-0.3cm}
\end{table}

\noindent\textbf{Mixed-domain head.}
Table~\ref{tab:headline} reports the main scene-disjoint comparison; MegaUnScene training scenes are seen in training, test scenes are not. 
The one-pass head matches the orbit teacher indoors (AUSE $0.17$) and improves over native aleatoric confidence OOD ($0.20$ vs.\ $0.32$), while trailing the $K{=}6$ teacher ($0.16$).
For refinement, the gate improves over feed-forward-only indoors ($53.5$ vs.\ $51.6$ AUC@30) and matches BA-always within $0.1$ AUC@30 ($53.5$ vs.\ $53.6$). 
OOD, the gate strongly avoids the collapse of BA-always ($58.4$ vs.\ $38.4$), but remains below FF-always ($60.3$). This indicates that, in the OOD setting, the main value of the gate is preventing harmful refinement rather than improving every feed-forward estimate.

\begin{table}[t]
\centering
\small
\caption{Scene-disjoint generalization of \method{}-O trained on 7-Scenes+MegaUnScene (mix) or on 7-Scenes only (7S) and evaluated on held-out indoor and outdoor scenes. AUSE $\downarrow$ measures uncertainty ranking; final AUC@30 $\uparrow$ is measured after the refinement decision. Best non-oracle deployable value per column is bold; teacher and oracle rows sit below the rule.}
\label{tab:headline}
\vspace{-0.2cm}
\resizebox{0.98\linewidth}{!}{
\begin{tabular}{lcccc}
\toprule
 & \multicolumn{2}{c}{Uncertainty AUSE $\downarrow$}
 & \multicolumn{2}{c}{Final AUC@30 $\uparrow$} \\
\cmidrule(lr){2-3}\cmidrule(lr){4-5}
Method & indoor & outdoor/OOD & indoor & outdoor/OOD \\
\midrule
Native aleatoric (VGGT)       & 0.19 & 0.32 & --   & --   \\
\textbf{\method{}-O (mix)}   & \textbf{0.17} & \textbf{0.20} & 53.5 & 58.4 \\
\method{}-O (7S), zero-shot & 0.21 & 0.21 & 53.0 & 47.8 \\
Feed-forward always           & --   & --   & 51.6 & \textbf{60.3} \\
BA always                     & --   & --   & \textbf{53.6} & 38.4 \\
\midrule
Orbit variance ($K{=}6$, teacher) & 0.17 & 0.16 & --   & --   \\
Oracle                        & --   & --   & 56.0 & 60.9 \\
\bottomrule
\end{tabular}}
\vspace{-0.3cm}
\end{table}

\noindent\textbf{Pose accuracy across thresholds.}
Indoors, the gate improves AUC@5/10/20 over feed-forward-only by $1.7$--$2.4$ points; OOD it remains $1.4$--$2.0$ points below feed-forward-only but $14.3$--$21.6$ points above BA-always. Thus, \method{} acts mainly as a control signal: it avoids harmful refinement while ranking poses by risk.

\subsection{Comparison with uncertainty baselines}
\label{sec:exp_uq}

We compare \method{} with uncertainty baselines on the same scene-disjoint splits.  \method{}-E, the error-supervised conditioning head, gives the best single-pass OOD AUSE and the best indoor AUSE among heads that do not read backbone-internal signals. The token-only head is competitive indoors but degrades OOD, while the parallax heuristic alone is insufficient in the scene-disjoint multi-view setting. Backbone-internal signals follow the same pattern: adding camera tokens ($0.132/0.224$) or attention statistics ($0.147/0.293$ alone, $0.118/0.204$ with conditioning) improves the indoor split but degrades OOD relative to the scalars-only head. We therefore keep the interpretable scalars as the default input.  Table~\ref{tab:uq} lists \method{}-E and the multi-task \method{}-O of Table~\ref{tab:headline} separately because their supervision targets differ.

\noindent\textbf{Per-feature ablation.}
Table~\ref{tab:ablation} retrains \method{}-O (mix) with one feature or feature group removed. The largest OOD degradation comes from removing the camera-head correction ($0.209\!\to\!0.273$), while the largest indoor degradation comes from removing native confidence ($0.165\!\to\!0.210$). The backbone stability signals thus carry most of the ranking: alone they reach $0.193/0.246$, against $0.365/0.359$ for the four geometric cues alone. Individual geometric scalars are partly redundant, but jointly they improve on the stability signals by $0.03$--$0.04$ AUSE, so the reliability signal combines backbone stability with geometric conditioning rather than resting on geometry alone.

We also compare against stronger epistemic-UQ baselines. A five-member deep ensemble~\cite{deepensembles} and MC-dropout~\cite{gal2016dropout} obtain OOD AUSE $0.20$ and $0.195$, respectively, comparable to our single-pass head but requiring roughly five backbone passes.

\begin{table}[t]
\centering
\small
\caption{Scene-disjoint uncertainty baselines. AUSE $\downarrow$ evaluates how well each score ranks pose error; the orbit ensemble is a training teacher, not an oracle.}
\label{tab:uq}
\vspace{-0.2cm}
\resizebox{0.75\linewidth}{!}{
\begin{tabular}{lcc}
\toprule
Uncertainty predictor & indoor & OOD \\
\midrule
Native aleatoric (VGGT)                  & 0.186 & 0.323 \\
Parallax heuristic                       & 0.756 & 0.560 \\
Learned: tokens-only                     & 0.152 & 0.227 \\
Learned: tokens+conditioning        & 0.132 & 0.224 \\
Learned: attention statistics       & 0.147 & 0.293 \\
Learned: conditioning+attention     & \textbf{0.118} & 0.204 \\
Learned: heteroscedastic-NLL             & 0.312 & 0.339 \\
\method{}-E (mix), single-output    & 0.124 & \textbf{0.196} \\
\method{}-O (mix) (Tab.~\ref{tab:headline}) & 0.168 & 0.204 \\
\midrule
Orbit variance ($K{=}6$, teacher)        & 0.173 & 0.163 \\
\bottomrule
\end{tabular}}
\end{table}

\begin{table}[t]
\centering
\small
\caption{Which features matter. The \method{}-O (mix) head is retrained in the protocol of Table~\ref{tab:headline} after removing one feature/group; mean$\pm$std over five seeds, $\Delta$ is the paired AUSE change against the full head, where positive is worse.}
\label{tab:ablation}
\vspace{-0.2cm}
\resizebox{0.98\linewidth}{!}{
\begin{tabular}{lcccc}
\toprule
  & \multicolumn{2}{c}{indoor AUSE $\downarrow$} 
  & \multicolumn{2}{c}{OOD AUSE $\downarrow$} \\
\cmidrule(lr){2-3}\cmidrule(lr){4-5}
Input set  & AUSE & $\Delta$ & AUSE & $\Delta$ \\
\midrule
all conditioning features (default) & 0.165$\pm$0.006 & +0.000 & 0.209$\pm$0.004 & +0.000 \\
\midrule
$-$ parallax & 0.149$\pm$0.003 & -0.016 & 0.213$\pm$0.003 & +0.003 \\
$-$ overlap / co-visibility & 0.157$\pm$0.003 & -0.008 & 0.215$\pm$0.003 & +0.005 \\
$-$ baseline & 0.161$\pm$0.006 & -0.004 & 0.202$\pm$0.004 & -0.007 \\
$-$ relative rotation & 0.172$\pm$0.008 & +0.007 & 0.200$\pm$0.004 & -0.009 \\
$-$ camera-head correction (rot.\ + trans.) & 0.191$\pm$0.003 & +0.026 & 0.273$\pm$0.005 & +0.064 \\
$-$ native confidence & 0.210$\pm$0.005 & +0.045 & 0.227$\pm$0.006 & +0.017 \\
native confidence only & 0.250$\pm$0.020 & +0.085& 0.339$\pm$0.003 & +0.130 \\
geometry only (parallax, overlap, baseline, rotation) & 0.365$\pm$0.016 & +0.200 & 0.359$\pm$0.009 & +0.149 \\
stability only (camera-head correction, native conf.) & 0.193$\pm$0.008 & +0.028 & 0.246$\pm$0.003 & +0.037 \\
\bottomrule
\end{tabular}}
\end{table}

\begin{table}[!t]
\centering
\small
\caption{Condensed adaptation results on independent MegaUnScene edge sets. Values are AUSE $\downarrow$. Cycle-distilled heads provide a ground-truth-free target-backbone adaptation route; cycle fusion uses raw cycle residuals at inference and is not a single-pass score; best single-pass value per row in bold; or VGGT, the VGGT-trained head is \method{}-E itself (parenthesized).
} 
\label{tab:adaptation}
\vspace{-0.2cm}
\resizebox{0.98\linewidth}{!}{
\begin{tabular}{lccccc}
\toprule
Backbone & Native & VGGT-trained & \method{}-E & \method{}-C & Cycle fusion \\
\midrule
VGGT        & 0.279 & (0.230) & \textbf{0.230} & 0.287 & 0.108 \\
$\pi^3$     & 0.333 & 0.286 & 0.221 & \textbf{0.187} & 0.108 \\
DUSt3R      & 0.182 & 0.460 & \textbf{0.162} & 0.211 & 0.119 \\
Fast3R      & 0.238 & 0.230 & \textbf{0.184} & 0.185 & 0.132 \\
CUT3R       & 0.391 & 0.334 & 0.314 & \textbf{0.305} & 0.150 \\
MapAnything & \textbf{0.209} & 0.496 & 0.240 & 0.251 & 0.147 \\
\bottomrule
\end{tabular}}
\vspace{-0.3cm}
\end{table}

\subsection{Adaptation across domains and backbones}
\label{sec:exp_adaptation}

The practical question is not whether one training recipe wins everywhere, but which adaptation setting should be used. We evaluate three settings on independent MegaUnScene edge sets: the VGGT-trained head applied without retraining, ground-truth-supervised target training (\method{}-E), and ground-truth-free cycle distillation (\method{}-C). 
Table~\ref{tab:adaptation} shows that no single source dominates all backbones and that gains are backbone-dependent: target supervision beats native confidence on five of six backbones, while the VGGT-trained head and cycle distillation help on $\pi^3$, Fast3R, and CUT3R but not on DUSt3R or MapAnything. Cycle-distilled heads provide a ground-truth-free adaptation route for some backbones, including $\pi^3$, where frame-permutation orbit variance vanishes because the model is permutation-equivariant. When independent cycles are available at inference, rank fusion with cycle residuals gives the strongest score, but this requires independent-edge graphs rather than a single joint pass.

\subsection{Reliability as a control signal}
\label{sec:exp_control}


\begin{table}[t]
\centering
\small
\caption{Reliability-driven control. Left: adaptive capture reports average frames needed to match all-8-frame accuracy. Right: pseudo-label yield at controlled catastrophic-label rate on OOD MegaUnScene.}
\label{tab:control}
\vspace{-0.2cm}
\resizebox{0.98\linewidth}{!}{
\begin{tabular}{lcc|ccccc}
\toprule
\multicolumn{3}{c|}{Adaptive capture} & \multicolumn{5}{c}{Pseudo-label curation} \\
Policy & strict & 5\% & $\alpha$ & Native & Ours & Oracle & risk \\
\midrule
Oracle & 2.10 & --   & 0.05 & 18\% & 38\% & 82\% & 0.047 \\
\method{} set-$u$ & 3.04 & 2.86 & 0.10 & 36\% & 64\% & 86\% & 0.095 \\
set-$u$ + native & 3.27 & 2.77 & 0.20 & 86\% & 92\% & 97\% & 0.197 \\
Native confidence & 3.85 & 3.33 & -- & -- & -- & -- & -- \\
Best fixed-$N$ & 4.00 & 3.00 & -- & -- & -- & -- & -- \\
Capture all 8 & 8.00 & 8.00 & -- & -- & -- & -- & -- \\
\bottomrule
\end{tabular}}
\vspace{-0.4cm}
\end{table}

\noindent\textbf{Calibration.}
We Platt-scale~\cite{platt1999probabilistic} $\hat{u}_{ij}$ into $\mathbb{P}(e_{ij}>\tau)$ on held-out splits. 
The calibrated \method{} head has low ECE ($0.061$ in-domain, $0.038$ OOD) and strong failure discrimination (AUROC $0.85$ and $0.80$). Native confidence can also be calibrated, but is less discriminative OOD, with AUROC dropping to $0.66$.

\noindent\textbf{Risk-controlled curation and adaptive capture.}
The calibrated reliability signal also supports threshold-based control decisions (Table~\ref{tab:control}). For pseudo-label curation, we keep labels whose calibrated uncertainty falls below a split-conformal threshold at a target catastrophic-label rate $\alpha$~\cite{vovk2005algorithmic,angelopoulos2023gentle,bates2021risk}. On MegaUnScene, at $\alpha{=}0.10$, \method{} keeps $64\%$ of pose pseudo-labels versus $36\%$ for native confidence at the same realised risk. The same signal can guide capture: stopping when a set-level aggregate of pairwise calibrated uncertainties (set-$u$ in Table~\ref{tab:control}) falls below a conformal threshold reaches the accuracy of using all eight frames with about $3.0$ frames on average, compared with $3.85$ frames for native-confidence stopping and $4.0$ frames for the best fixed-$N$ policy.

\noindent\textbf{Pose-graph fusion.}
We test whether predicted uncertainty helps when independent pairwise relative poses are fused by rotation averaging. 
On MegaUnScene, \method{} reduces median global rotation error from $15.1^\circ\,[12.5,17.5]$ with uniform weighting to $12.7^\circ\,[10.1,15.1]$ with uncertainty weighting and $11.6^\circ\,[9.4,14.1]$ with an uncertainty-selected tree ($N=6$, $80$ view sets; full plot in the supplementary material). These outperform random-tree selection ($14.8^\circ$) and native-confidence weighting ($13.9^\circ$), with oracle at $9.6^\circ$. On held-out 7-Scenes, gains are smaller and not statistically significant, as expected when catastrophic edges are less frequent.

%% file: sec/5_discussion_limitations.tex
\section{Discussion and limitations}
\label{sec:discussion} 

Our results suggest that reliability in feed-forward 3D reconstruction is not only a matter of improving the backbone, but also of estimating when its predictions should be trusted. \method{} does not modify the backbone or directly improve every feed-forward pose estimate. Instead, it makes the output usable: it predicts which relative poses are reliable, when refinement should be applied, which edges should be down-weighted or pruned, and when additional capture is necessary.  The direct pose-accuracy benefit is therefore setting-dependent: the gate improves accuracy when refinement often helps, and mainly prevents harmful refinement when it does not. It also needs labelled pairs from a domain where BA behaves as in the target; otherwise, keeping the feed-forward pose is the safer policy.

The one-pass head is not a claim that distillation always beats its teacher: the orbit-variance ensemble remains stronger when its $K$-pass budget is available. The head's advantages are deployment cost, compatibility with downstream decisions, and adaptability through other teachers. In particular, cycle-residual distillation provides a ground-truth-free adaptation route for the uncertainty output and works for equivariant models where permutation variance vanishes.

The signal helps most where native confidence is blind, such as indoor low-parallax and low-overlap cases and outdoor extreme-view data with large relative rotations. It may require target-domain supervision or calibration when native confidence already tracks the dominant failure mode. Absolute failure-probability calibration is also domain-specific; ranking transfers more robustly than the probability scale. Finally, full downstream SfM/SLAM trajectory-error evaluation remains future work.

%% file: sec/6_conclusion.tex
\section{Conclusion}
\label{sec:conclusion}

Feed-forward 3D models are fast enough to become building blocks, but need a trust interface. 
We presented \method{}, a conditioning-aware reliability adapter for frozen feed-forward 3D backbones. \method{} reads backbone-predicted geometry, estimates pose-level reliability, and converts it into decisions about refinement, pruning, calibration, capture, and pose-graph fusion.
On VGGT, \method{} distils a training-time orbit-variance teacher into a single-pass uncertainty predictor and learns when to invoke BA where labelled training pairs exist. It improves pose-error ranking over native confidence, also zero-shot from indoor to outdoor scenes, supports ground-truth-free cycle-residual adaptation for some backbones, and makes downstream geometric estimation more robust. Feed-forward 3D systems should expose not only geometry, but also reliability over that geometry.

%% file: sec/supp.tex
\clearpage
\hypersetup{pageanchor=false}
\setcounter{page}{1}
\maketitlesupplementary


\vspace{-6pt}
\begin{strip}
\centering
\includegraphics[width=0.95\textwidth]{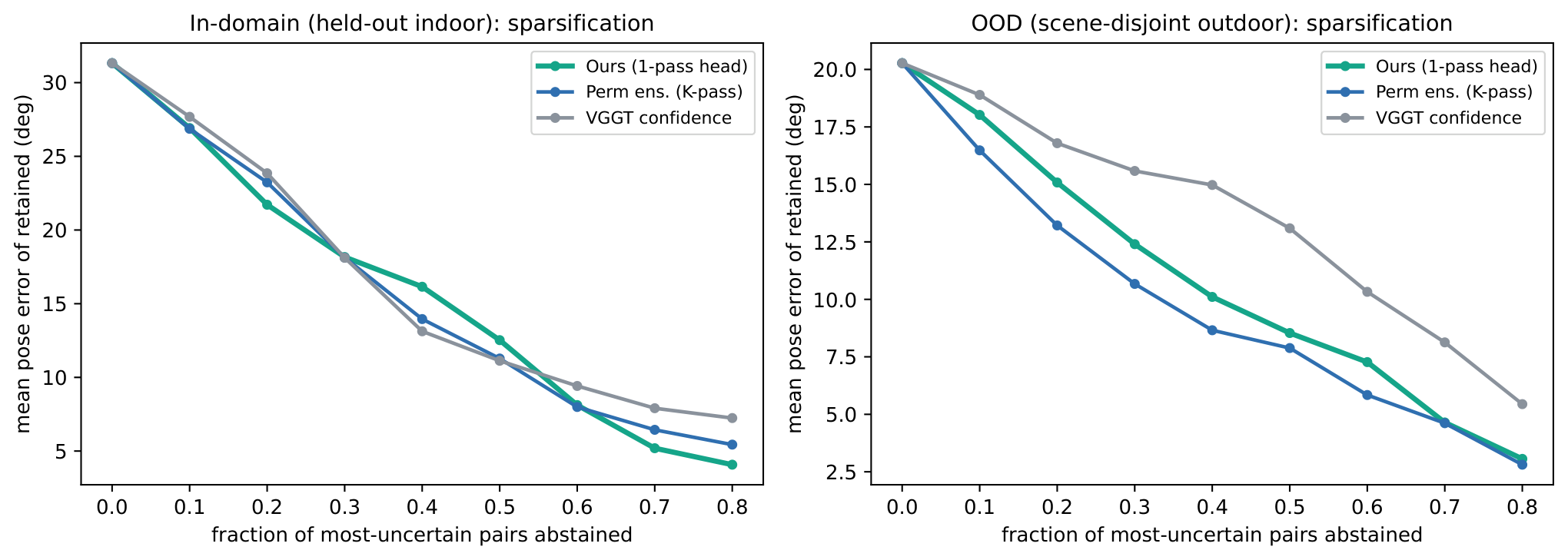}
\vspace{-0.3cm}
\captionof{figure}{
Sparsification curves: mean pose error of retained pairs after abstaining from the most uncertain predictions. The one-pass head and the orbit teacher track pose error substantially better than native confidence.
}
\label{fig:supp_spars}
\end{strip}

\section{Sparsification and two-view uncertainty}
\label{supp:uq}
Figure~\ref{fig:supp_spars} shows the operational meaning of AUSE: when the most uncertain pairs are removed, retained-pair error decreases substantially faster for the one-pass head and the orbit teacher than for native confidence.
Table~\ref{tab:supp_twoview} shows that on standard two-view benchmarks, simple geometric cues such as predicted parallax and orbit disagreement are already strong uncertainty signals. This supports the interpretation that \method{} is most useful when multiple conditioning cues must be combined, as in multi-view and extreme-view settings.

\begin{table}[h]
\centering
\small
\caption{Uncertainty predictors on standard two-view benchmarks, measured by AUSE $\downarrow$. Predicted parallax and orbit disagreement are strong cues on these pairs; \method{} remains competitive, with its main advantage in multi-view and extreme-view settings.}
\label{tab:supp_twoview}
\vspace{-0.2cm}
\resizebox{0.98\linewidth}{!}{
\begin{tabular}{lccccc}
\toprule
Benchmark 
& Native 
& RANSAC-inlier 
& Pred.\ parallax 
& Orbit 
& Ours \\
\midrule
MegaDepth-1500 ($n{=}1500$) & 0.616 & 0.874 & 0.245 & 0.190 & 0.261 \\
ScanNet-1500 ($n{=}1500$)   & 0.386 & 0.832 & 0.202 & 0.181 & 0.230 \\
\bottomrule
\end{tabular}}
\vspace{-0.3cm}
\end{table}

\section{Calibration of catastrophic-pose failure}
\label{supp:calibration}

Table~\ref{tab:supp_calibration} reports post-hoc calibration for catastrophic error. Native confidence achieves low ECE after calibration, but its OOD AUROC is substantially weaker. This supports the main-paper conclusion that \method{} is more useful for OOD control decisions because it remains discriminative after calibration.

\begin{table}[h]
\centering
\small
\caption{Post-hoc calibration of catastrophic-failure probability $\mathbb{P}(e_{ij}>30^\circ)$. Native confidence can be calibrated but is less discriminative OOD.}
\label{tab:supp_calibration}
\vspace{-0.2cm}
\resizebox{0.98\linewidth}{!}{
\begin{tabular}{lcccccc}
\toprule
 & \multicolumn{3}{c}{Indoor} & \multicolumn{3}{c}{OOD} \\
\cmidrule(lr){2-4}\cmidrule(lr){5-7}
Score & ECE $\downarrow$ & Brier $\downarrow$ & AUROC $\uparrow$ & ECE $\downarrow$ & Brier $\downarrow$ & AUROC $\uparrow$ \\
\midrule
Native confidence & \textbf{0.034} & 0.146 & 0.83 & \textbf{0.008} & 0.163 & 0.66 \\
\textbf{\method{}} & 0.061 & \textbf{0.143} & \textbf{0.85} & 0.038 & \textbf{0.142} & \textbf{0.80} \\
\bottomrule
\end{tabular}}
\end{table}

\section{Downstream pose-graph fusion}
\label{supp:posegraph}

\begin{figure*}[!t]
\centering
\includegraphics[width=0.9\linewidth]{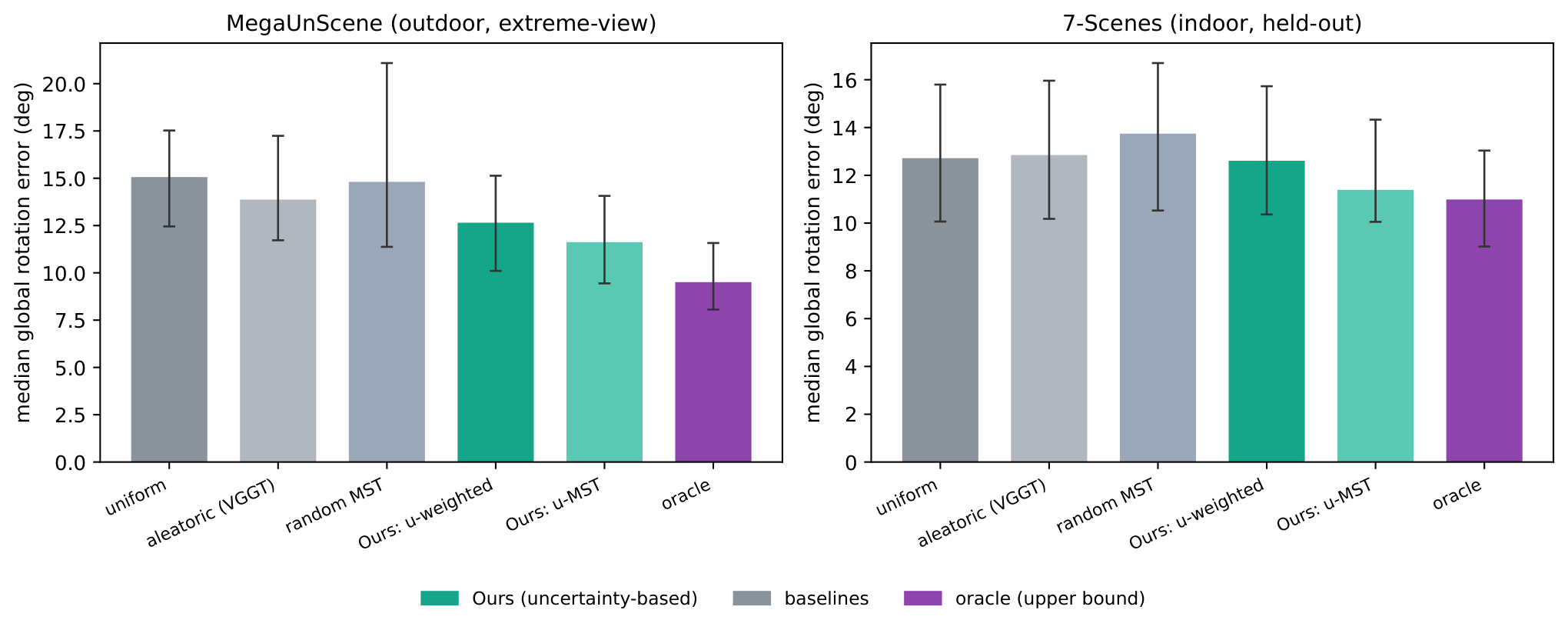}
\vspace{-0.3cm}
\caption{Downstream pose-graph fusion from independent pairwise VGGT estimates ($N=6$, $80$ view sets). Bars show median global rotation error with bootstrap $95\%$ CIs. Uncertainty-weighted and uncertainty-selected edges improve rotation averaging on MegaUnScene; gains are smaller indoors.}
\label{fig:supp_downstream}
\end{figure*}

Figure~\ref{fig:supp_downstream} visualizes the downstream rotation-averaging experiment discussed in main-paper Sec. 4.5. The figure reports median global rotation error with bootstrap $95\%$ confidence intervals for independent pairwise VGGT estimates.

\section{Streaming frame reduction}
\label{supp:streaming}

For a bounded memory stream, we keep a fixed-size buffer and evict the highest-uncertainty frame when a new frame arrives. Table~\ref{tab:supp_streaming} shows that uncertainty-driven eviction tracks the offline ceiling on hard Co3D object orbits, especially under high discard rates.

\begin{table}[!t]
\centering
\small
\caption{Streaming frame reduction on hard Co3D object orbits. Entries report median pose error in degrees, with largest uncovered gap in parentheses. Lower is better for both.}
\label{tab:supp_streaming}
\vspace{-0.2cm}
\resizebox{0.98\linewidth}{!}{
\begin{tabular}{lccc}
\toprule
Policy & 50\% discard & 70\% discard & 90\% discard \\
\midrule
Full set (no discard) & -- & 67.8 (0.00) & -- \\
Uniform every $k$-th & 2.8 (0.11) & 67.1 (0.19) & 60.1 (0.49) \\
Reservoir random & 67.3 (0.21) & 66.3 (0.29) & 53.2 (0.65) \\
Overlap heuristic & 1.3 (0.12) & 1.3 (0.27) & 31.9 (0.72) \\
VGGT aleatoric & 1.1 (0.11) & 1.3 (0.48) & 2.6 (0.81) \\
$u \circ$ aleatoric fusion & 1.2 (0.11) & 1.3 (0.43) & 1.8 (0.66) \\
\method{} $u$-eviction & 1.1 (0.11) & 1.3 (0.28) & 1.8 (0.58) \\
Batch min-$u$ offline ceiling & 1.2 (0.11) & 1.2 (0.27) & 1.6 (0.56) \\
\bottomrule
\end{tabular}}
\vspace{-0.3cm}
\end{table}

\section{Cheap and noisy backbone variants}
\label{supp:cheap}

\begin{table*}[!t]
\centering
\small
\caption{Cheap-resolution and weak-variant rescue. We report AUC@30 for all pairs and for the kept half after pruning. Head-pruning uses the \method{} reliability score and is the deployed selector; bold marks the recommended selector for the weak-variant rows. For VGGT resolution rows, values are MegaDepth / ScanNet.}
\label{tab:supp_cheap}
\vspace{-0.2cm}
\resizebox{0.75\linewidth}{!}{
\begin{tabular}{lcccccc}
\toprule
Variant / resolution & ms/pair & base/ref & all & random 50\% & native 50\% & head 50\% \\
\midrule
VGGT 224px (MD/SN) & 51 & -- & 53.8 / 43.7 & 53.8 / 44.1 & 57.3 / 46.4 & 60.7 / 50.5 \\
VGGT 364px (MD/SN) & 55 & -- & 73.8 / 74.4 & 73.8 / 74.5 & 73.7 / 76.4 & 81.0 / 79.6 \\
VGGT 518px (MD/SN) & 70 & -- & 82.3 / 83.1 & 82.1 / 83.5 & 82.5 / 86.0 & 90.4 / 91.3 \\
StreamVGGT & 75 & 69.5 & 60.8 & 60.0 & 68.5 & \textbf{77.7} \\
QuantVGGT W4A4 & 30 & 69.5 & 63.8 & 63.1 & 75.0 & \textbf{82.7} \\
DUSt3R 224-linear & 104 & 47.9 & 44.6 & 44.9 & 65.0 & \textbf{70.8} \\
CUT3R 224-linear & 64 & 55.0 & 44.7 & 44.7 & 60.2 & \textbf{64.1} \\
\bottomrule
\end{tabular}}
\vspace{-0.3cm}
\end{table*}

\begin{table*}[t]
\centering
\small
\caption{Relative-pose AUC at several thresholds for the refinement policies of main-paper Table~3. The same pairs are used as in the main headline table; gate denotes the seed-0 deployed head, with the five-seed mean shown below. The oracle row is included as an upper reference, and the best non-oracle value per column is bold.}
\label{tab:auc}
\vspace{-0.2cm}
\resizebox{0.8\linewidth}{!}{
\begin{tabular}{lcccccccc}
\toprule
 & \multicolumn{4}{c}{indoor (7-Scenes office)} & \multicolumn{4}{c}{outdoor/OOD (MegaUnScene)} \\
\cmidrule(lr){2-5}\cmidrule(lr){6-9}
Policy & @5 & @10 & @20 & @30 & @5 & @10 & @20 & @30 \\
\midrule
Feed-forward always & 14.1& 26.5 & 43.0 & 51.6 & \textbf{25.1} & \textbf{39.7} & \textbf{53.0} & \textbf{60.3} \\
BA always & \textbf{16.8} & \textbf{30.0} & \textbf{45.0} & \textbf{53.6} & 9.4 & 17.5 & 29.5 & 38.4 \\
\textbf{\method{}-O (mix) gate} & 15.8 & 28.9 & 44.8 & 53.5 & 23.7 & 38.0 & 51.0 & 58.4 \\
\quad5-seed mean & 15.7$\pm$0.1 & 28.7$\pm$0.1 & 44.6$\pm$0.1 & 53.3$\pm$0.1 & 23.1$\pm$0.3 & 37.2$\pm$0.4 & 50.1$\pm$0.5 & 57.6$\pm$0.5 \\
\method{}-O (7S) gate, 5-seed mean & 16.3$\pm$0.2 & 29.4$\pm$0.2 & 44.8$\pm$0.3 & 53.4$\pm$0.3 & 13.8$\pm$0.6 & 24.1$\pm$1.0 & 37.1$\pm$1.4 & 46.0$\pm$1.5 \\
\midrule
Oracle gate & 18.9 & 32.5 & 48.0 & 56.0 & 25.9 & 40.3 & 53.6 & 60.9 \\
\bottomrule
\end{tabular}}
\vspace{-0.3cm}
\end{table*}

The same reliability signal can trade compute for quality. Table~\ref{tab:supp_cheap} shows that reducing VGGT input resolution hurts raw accuracy, but uncertainty pruning recovers much of the loss. A $364$-pixel pass with head-pruning approaches the full $518$-pixel pass without pruning ($81.0$ vs.\ $82.3$ AUC@30 on MegaDepth at $50\%$ kept; $79.6$ vs.\ $83.1$ on ScanNet). The same trend holds for weaker or noisy variants such as StreamVGGT, QuantVGGT, DUSt3R, and CUT3R.

\section{Feed-forward residual}
\label{supp:residual}
We also test whether a single-pass SE(3) residual can replace inference-time BA. A naive residual is worse than the feed-forward pose; a small delta-from-FF residual is safe but does not recover BA's gains (office $51.8$ vs.\ $51.6$ AUC@30; OOD $60.1$ vs.\ $60.3$). Thus, the practical value comes from deciding \emph{when} to refine rather than amortizing BA itself.

\section{Repeated-seed stability}
\label{supp:seeds}
Across five repeated-seed runs, the scene-disjoint general head remains stable: uncertainty AUSE is $0.165{\pm}0.006$ indoors and $0.209{\pm}0.004$ OOD, while final AUC@30 after gating is $53.3{\pm}0.1$ indoors and $57.6{\pm}0.5$ OOD. These values are consistent with the single-run headline table in the main paper.

\section{Pose accuracy across thresholds}
\label{supp:auc}

Table~\ref{tab:auc} lists the AUC@5/10/20/30 values behind the threshold analysis of main-paper Sec. 4.2. At every threshold, the mixed-domain gate improves on feed-forward-only indoors and, OOD, sits between feed-forward-only and BA-always, far above the latter. The strict zero-shot gate (7S), trained on 7-Scenes only, stays well below feed-forward-only OOD at every threshold. Two refine-or-keep rules that need no target-domain labels, refining when the predicted uncertainty is below, or the predicted overlap above, a threshold chosen on the 7-Scenes training pairs, reach $38.8$ and $44.1$ OOD AUC@30 (five seeds), below the transferred gate ($46.0$). The gate therefore needs labelled pairs from a domain in which BA behaves as in the target, whereas the uncertainty output transfers without target labels.

\section{Seed robustness of the learned heads}
\label{supp:seeds_uq}

Table~\ref{tab:supp_uq_seeds} extends the seed analysis to every learned row of main-paper Table~4, under both supervision targets. The seed spread is $0.002$--$0.03$ AUSE and the ranking of the rows is unchanged, showing that the distinction between \method{}-E (main-paper Table~4) and the multi-task \method{}-O (main-paper Table~3) is seed-stable. The lower block retrains the backbone-internal variants in the multi-task protocol of main-paper Table~3; every internal-signal variant is worse than the conditioning head on both splits, and the multi-layer camera tokens are unstable across seeds, consistent with the token-overfitting observation in main-paper Sec.~4.2.

\begin{table*}[t]
\centering
\small
\caption{Five-seed mean$\pm$std of the learned heads on the pairs of main-paper Table~4, under both supervision targets, and of heads reading backbone-internal signals in the multi-task protocol of main-paper Table~3.}
\label{tab:supp_uq_seeds}
\vspace{-0.2cm}
\resizebox{0.8\textwidth}{!}{
\begin{tabular}{lcccc}
\toprule
 & \multicolumn{2}{c}{AUSE $\downarrow$} & \multicolumn{2}{c}{Spearman $\uparrow$} \\
\cmidrule(lr){2-3}\cmidrule(lr){4-5}
Uncertainty predictor & indoor & OOD & indoor & OOD \\
\midrule
Native aleatoric (VGGT) & 0.186$\pm$0.000 & 0.323$\pm$0.000 & 0.59$\pm$0.00 & 0.46$\pm$0.00 \\
Parallax heuristic & 0.756$\pm$0.000 & 0.560$\pm$0.000 & 0.09$\pm$0.00 & 0.17$\pm$0.00 \\
Learned: tokens-only (error-sup.) & 0.150$\pm$0.012 & 0.234$\pm$0.017 & 0.70$\pm$0.02 & 0.61$\pm$0.02 \\
Learned: tokens-only (orbit-distilled) & 0.220$\pm$0.029 & 0.269$\pm$0.009 & 0.63$\pm$0.02 & 0.57$\pm$0.02 \\
Learned: tokens+conditioning (error-sup.) & 0.150$\pm$0.012 & 0.228$\pm$0.011 & 0.71$\pm$0.02 & 0.62$\pm$0.02 \\
Learned: tokens+conditioning (orbit-distilled) & 0.190$\pm$0.011 & 0.260$\pm$0.022 & 0.65$\pm$0.01 & 0.60$\pm$0.02 \\
Learned: heteroscedastic-NLL (error-sup.) & 0.328$\pm$0.024 & 0.364$\pm$0.018 & 0.50$\pm$0.03 & 0.46$\pm$0.02 \\
Learned: attention/feature statistics (error-sup.) & 0.144$\pm$0.006 & 0.299$\pm$0.021 & 0.69$\pm$0.01 & 0.53$\pm$0.02 \\
Learned: attention/feature statistics (orbit-distilled) & 0.199$\pm$0.018 & 0.319$\pm$0.014 & 0.61$\pm$0.02 & 0.47$\pm$0.01 \\
Learned: conditioning+attention stats (error-sup.) & 0.107$\pm$0.009 & 0.213$\pm$0.010 & 0.78$\pm$0.01 & 0.69$\pm$0.01 \\
Learned: conditioning+attention stats (orbit-distilled) & 0.155$\pm$0.011 & 0.245$\pm$0.011 & 0.71$\pm$0.01 & 0.60$\pm$0.02 \\
\textbf{\method{}-E (mix)}& 0.125$\pm$0.005 & 0.198$\pm$0.002 & 0.80$\pm$0.00 & 0.68$\pm$0.00 \\
\textbf{\method{}-O (mix), single-output} & 0.163$\pm$0.007 & 0.199$\pm$0.002 & 0.73$\pm$0.01 & 0.67$\pm$0.00 \\
\textbf{\method{}-O (mix), multi-task (main-paper Table~3)} & 0.165$\pm$0.006 & 0.209$\pm$0.004 & 0.72$\pm$0.01 & 0.66$\pm$0.00 \\
\midrule
Orbit variance ($K{=}6$, teacher) & 0.173$\pm$0.000 & 0.163$\pm$0.000 & 0.72$\pm$0.00 & 0.70$\pm$0.00 \\
\midrule
\multicolumn{5}{l}{\emph{Backbone-internal signals, main-paper Table~3 protocol}} \\
conditioning (default head) & 0.165$\pm$0.006 & 0.209$\pm$0.004 & 0.72$\pm$0.01 & 0.66$\pm$0.00 \\
camera tokens only & 0.235$\pm$0.028 & 0.247$\pm$0.014 & 0.60$\pm$0.03 & 0.60$\pm$0.01 \\
conditioning + camera tokens & 0.219$\pm$0.015 & 0.252$\pm$0.018 & 0.61$\pm$0.02 & 0.59$\pm$0.01 \\
attention / feature statistics only & 0.206$\pm$0.016 & 0.313$\pm$0.008 & 0.60$\pm$0.02 & 0.50$\pm$0.02 \\
conditioning + attention / feature statistics & 0.150$\pm$0.007 & 0.231$\pm$0.008 & 0.71$\pm$0.01 & 0.63$\pm$0.01 \\
multi-layer camera tokens only & 0.287$\pm$0.141 & 0.318$\pm$0.020 & 0.54$\pm$0.15 & 0.51$\pm$0.02 \\
conditioning + multi-layer camera tokens & 0.313$\pm$0.113 & 0.307$\pm$0.023 & 0.51$\pm$0.11 & 0.52$\pm$0.03 \\
relative pose $\psi$ only & 0.621$\pm$0.033 & 0.640$\pm$0.025 & 0.14$\pm$0.04 & 0.01$\pm$0.01 \\
\bottomrule
\end{tabular}}
\end{table*}